\documentclass{sig-alternate}

\usepackage{url}
\usepackage{pgfplots}

\begin{document}

\permission{Permission to make digital or hard copies of part or all of this work for personal or classroom use is granted without fee provided that copies are not made or distributed for profit or commercial advantage and that copies bear this notice and the full citation on the first page. Copyrights for third-party components of this work must be honored. For all other uses, contact the Owner/Author. Copyright is held by the owner/author(s).}
\conferenceinfo{K-CAP 2015,}{October 07-10, 2015, Palisades, NY, USA}
\copyrightetc{ACM \the\acmcopyr}
\crdata{978-1-4503-3849-3/15/10\\
http://dx.doi.org/10.1145/2815833.2816946}

\title{Evaluating the Competency of a First-Order Ontology}

\numberofauthors{3}

\author{
\alignauthor
Javier \'Alvez\\
       \affaddr{LoRea Group}\\
       \affaddr{University of the Basque Country UPV/EHU}\\
       \email{javier.alvez@ehu.eus}
\alignauthor
Paqui Lucio\\
       \affaddr{LoRea Group}\\
       \affaddr{University of the Basque Country UPV/EHU}\\
       \email{paqui.lucio@ehu.eus}
\alignauthor German Rigau\\
       \affaddr{IXA NLP Group}\\
       \affaddr{University of the Basque Country UPV/EHU}\\
       \email{german.rigau@ehu.eus}
}

\date{\today}

\maketitle


\newcommand{\tab}{\hspace{0.2cm}}
\newcommand{\connective}[1]{\bf #1 \;}
\newcommand{\predicate}[1]{\rm #1}
\newcommand{\constant}[1]{\rm #1}
\newcommand{\variable}[1]{\tt ?#1}

\newcommand{\textConstant}[1]{{\it{#1}}}
\newcommand{\textPredicate}[1]{{\it{#1}}}

\newcommand{\ADIMENSUMO}{Adimen-SUMO }
\newcommand{\ADIMENSUMOWWS}{Adimen-SUMO}

\newcommand{\SUMO}{SUMO }
\newcommand{\SUMOWWS}{SUMO}

\newcommand{\WORDNET}{WordNet }
\newcommand{\WORDNETWWS}{WordNet}

\begin{abstract}

We report on the results of evaluating the competency of a first-order ontology for its use with automated theorem provers (ATPs). The evaluation follows the adaptation of the methodology based on competency questions (CQs) \cite{GrF95} to the framework of first-order logic, which is presented in \cite{ALR15}, and is applied to \ADIMENSUMO \cite{ALR12}. The set of CQs used for this evaluation has been automatically generated from a small set of semantic patterns and the mapping of \WORDNET to \SUMOWWS. Analysing the results, we can conclude that it is feasible to use ATPs for working with \ADIMENSUMO v2.4, enabling the resolution of goals by means of performing non-trivial inferences.

\end{abstract}



\section{Introduction}

Ontologies can be evaluated by considering their use in an application when performing correct predictions on inferencing \cite{PoM04}. In order to enable better reasoning capabilities, the inferencing process should be able to deduce from the ontology as much correct implicit knowledge as possible. In \cite{GrF95}, the authors propose to use a set of competency questions (CQs) to evaluate an ontology, which are goals that the ontology is expected to answer. Once the set of CQs is defined, the evaluation consists in checking whether the CQs are entailed by the ontology or not. 

State-of-the-art automatic theorem provers (ATP) for first-order logic (FOL) like Vampire \cite{RiV02} or E \cite{Sch02} are highly sophisticated systems that have been proved to provide advanced reasoning support to expressive ontologies. Consequently, these tools can be used to automatize the evaluation of first-order (FO) ontologies as proposed in \cite{GrF95}. More concretely, the authors of \cite{ALR15} adapt the above method for evaluating FO ontologies using FOL ATPs, and they also obtain semi-automatically a set of CQs from the mapping from \WORDNET (WN) to \SUMO \cite{Niles+Pease'03}.

In this paper, we successfully apply the method described in \cite{ALR15} for evaluating \ADIMENSUMO v2.4, which is a FO ontology obtained by transforming \SUMOWWS, using the proposed set of CQs. Our results prove that it is feasible to work with FO ontologies and FOL ATPs to perform non-trivial inferences that could be very useful for a wide range of knowledge intensive applications. For instance, to help validating the consistency of associated semantic resources like \WORDNETWWS, or to derive new explicit knowledge from them.

In the next two sections, we introduce \ADIMENSUMO and the method used in Section \ref{section:Evaluation} to evaluate the ontology. In the last section, we provide some concluding remarks.

\section{\ADIMENSUMOWWS} \label{section:AdimenSUMO}

\begin{table}[t]
\caption{\label{table:AdimenSUMOFigures} Some figures about \ADIMENSUMOWWS}
\centering
\begin{tabular} {lrr}
\hline \\[-6pt]
 &{\bf Unit clauses} & {\bf General clauses}  \\
\hline\rule{-4pt}{10pt}
Meta-knowledge & 0 & 8 \\
Top level of \SUMO & 2,034 & 650 \\
Mid level of \SUMO & 2,601 & 975 \\
FO transformation & 0 & 1,152 \\
{\bf Total} & {\bf 4,635} & {\bf 2,785} \\
\hline
\end{tabular}
\end{table}

\SUMOWWS\footnote{\url{http://www.ontologyportal.org}} \cite{Niles+Pease'01} has its origins in the nineties, when a group of engineers from the IEEE Standard Upper Ontology Working Group pushed for a formal ontology standard. Their goal was to develop a standard upper ontology to promote data interoperability, information search and retrieval, automated inference and natural language processing.

\SUMO is expressed in SUO-KIF (Standard Upper Ontology Knowledge Interchange Format \cite{Pea09}), which is a dialect of KIF (Knowledge Interchange Format \cite{Richard+'92}). Both KIF and SUO-KIF can be used to write FOL formulas, but its syntax goes beyond FOL. Consequently, \SUMO cannot be directly used by FOL ATPs without a suitable transformation. \ADIMENSUMO is the result of transforming around \%88 of axioms from the top and the middle levels of \SUMO into FOL formulas \cite{ALR12}. The translation is based on a small set of axioms, which provide the axiomatization of the meta-predicates that are required to define the knowledge of \SUMO as FOL formulas. Some of these meta-predicates are \textPredicate{instance}, \textPredicate{subclass}, \textPredicate{disjoint} and \textPredicate{partition}, whose axiomatization cannot be directly inherited from \SUMOWWS. The transformation also adds new axioms for a suitable characterization of \SUMO types, variable-arity relations and \textPredicate{holds$_k$} predicates, which simulate the use of variable-predicates in FOL formulas. In Table \ref{table:AdimenSUMOFigures}, we provide some figures about the content of \ADIMENSUMOWWS. More specifically, the number of atomic formulas (unit clauses) and non-atomic formulas (general clauses) that is used in each part of \ADIMENSUMOWWS. Roughly speaking, atomic formulas (or simply atoms) make the explicit knowledge of the ontology, whereas non-atomic formulas (which contain connectives and/or quantifiers) define the implicit knowledge. It is worth to note that the axiomatization of meta-predicates (meta-knowledge) and the axioms required for the transformation into FOL formulas (FO transformation) do not include any atomic formula. To sum up, we provide 4,635 unit clauses and 2,785 general clauses  plus a conjecture as source to the ATP for each CQ.

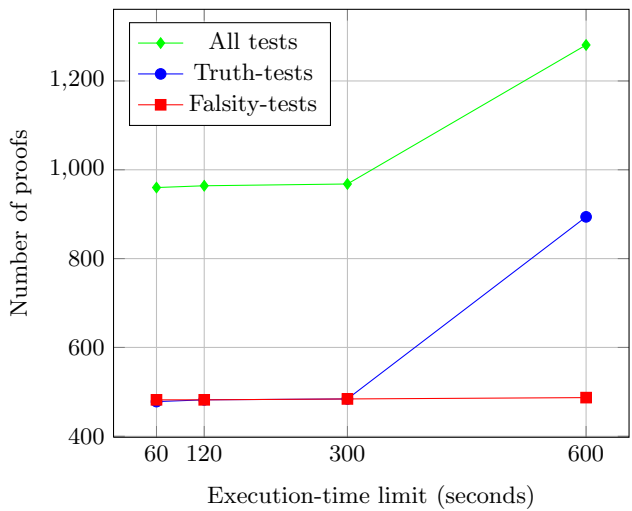
\begin{figure}[t]
\centering
\begin{tikzpicture}
\begin{axis}[
legend pos = north west,
xlabel = {Execution-time limit (seconds)},
ylabel = {Number of proofs},
xtick=data,
grid=major
]
\addplot[color=green,mark=diamond*] coordinates {
	(60,960)
	(120,964)
	(300,968)
	(600,1281)
};
\addplot[color=blue,mark=*] coordinates {
	(60,478)
	(120,482)
	(300,484)
	(600,894)
};
\addplot[color=red,mark=square*] coordinates {
	(60,482)
	(120,482)
	(300,484)
	(600,487)
};
\legend{ All tests, Truth-tests, Falsity-tests}
\end{axis}
\end{tikzpicture}
\caption{Number of solved goals}
\label{fig:SolvedGoals}
\end{figure}

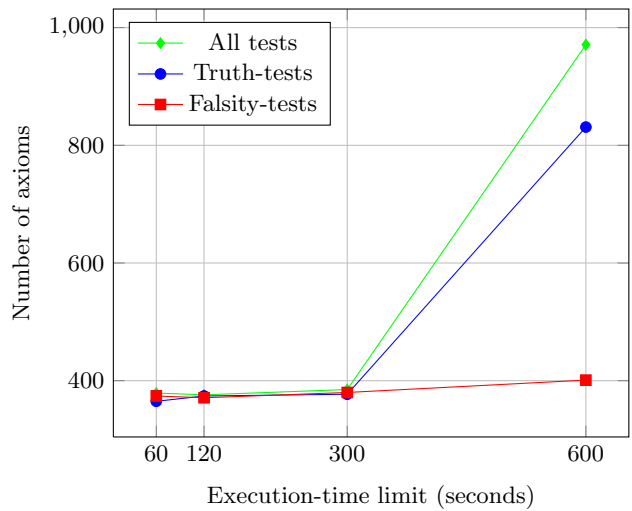
\begin{figure}[t]
\centering
\begin{tikzpicture}
\begin{axis}[
legend pos = north west,
xlabel = {Execution-time limit (seconds)},
ylabel = {Number of axioms},
xtick=data,
grid=major
]
\addplot[color=green,mark=diamond*] coordinates {
	(60,379)
	(120,376)
	(300,385)
	(600,971)
};
\addplot[color=blue,mark=*] coordinates {
	(60,365)
	(120,374)
	(300,377)
	(600,831)
};
\addplot[color=red,mark=square*] coordinates {
	(60,374)
	(120,371)
	(300,380)
	(600,401)
};
\legend{ All tests, Truth-tests, Falsity-tests}
\end{axis}
\end{tikzpicture}
\caption{Number of different axioms used in proofs}
\label{fig:TotalAxioms}
\end{figure}

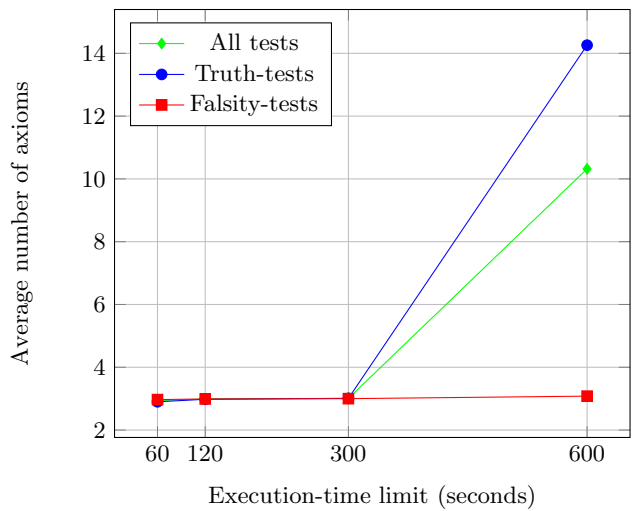
\begin{figure}
\centering
\begin{tikzpicture}
\begin{axis}[
legend pos = north west,
xlabel = {Execution-time limit (seconds)},
ylabel = {Average number of axioms},
xtick=data,
grid=major
]
\addplot[color=green,mark=diamond*] coordinates {
	(60,2.94)
	(120,2.99)
	(300,3.01)
	(600,10.32)
};
\addplot[color=blue,mark=*] coordinates {
	(60,2.9)
	(120,2.98)
	(300,3.01)
	(600,14.26)
};
\addplot[color=red,mark=square*] coordinates {
	(60,2.97)
	(120,2.99)
	(300,3)
	(600,3.08)
};
\legend{ All tests, Truth-tests, Falsity-tests}
\end{axis}
\end{tikzpicture}
\caption{Average number of axioms used in proofs}
\label{fig:AverageAxioms}
\end{figure}

\section{Evaluation Method} \label{section:EvaluationMethod}

In \cite{GrF95}, the authors propose to evaluate the expressiveness of an ontology by proving completeness theorems w.r.t. a set of CQs. The proof of completeness theorems requires to check whether a given CQ is entailed by the ontology or not. In \cite{ALR15}, the authors adapt the above method to use FOL ATPs as a tool for automatically checking whether a CQ is entailed or not. In particular, this adaptation is based on the use of FOL ATPs that work by refutation within a given execution-time limit. Following this proposal, the set of CQs is partitioned into two classes: {\it truth-tests} and {\it falsity-tests},  depending on whether we expect the conjecture to be entailed by the ontology or not. If the ATP is able to find a prove for a given conjecture, then we know for sure that the corresponding CQ is entailed by the ontology. However, if the ATP cannot find a proof, we do not know if the conjecture is not entailed by the ontology or although the conjecture is entailed, the ATP has not been able to find the proof within the provided execution-time limit. Consequently, and according to the proposal in \cite{ALR15}, a truth-test is classified as (i) {\it passing} if the ATP finds a proof, since the conjecture is expected to be entailed, whereas a falsity-tests is classified as (ii) {\it non-passing} if a proof is found, because the conjecture is expected not to be entailed. Otherwise, if no proof is found, both truth-tests and falsity tests are classified as (iii) {\it unknown}. This classification provides an effective method to evaluate an ontology w.r.t. a given set of CQs. It is worth to remark that the evaluation results strongly depend on the execution-time limit of the ATP, since a CQ classified as unknown within a given execution-time limit may be classified as passing/non-passing within a longer execution-time limit.

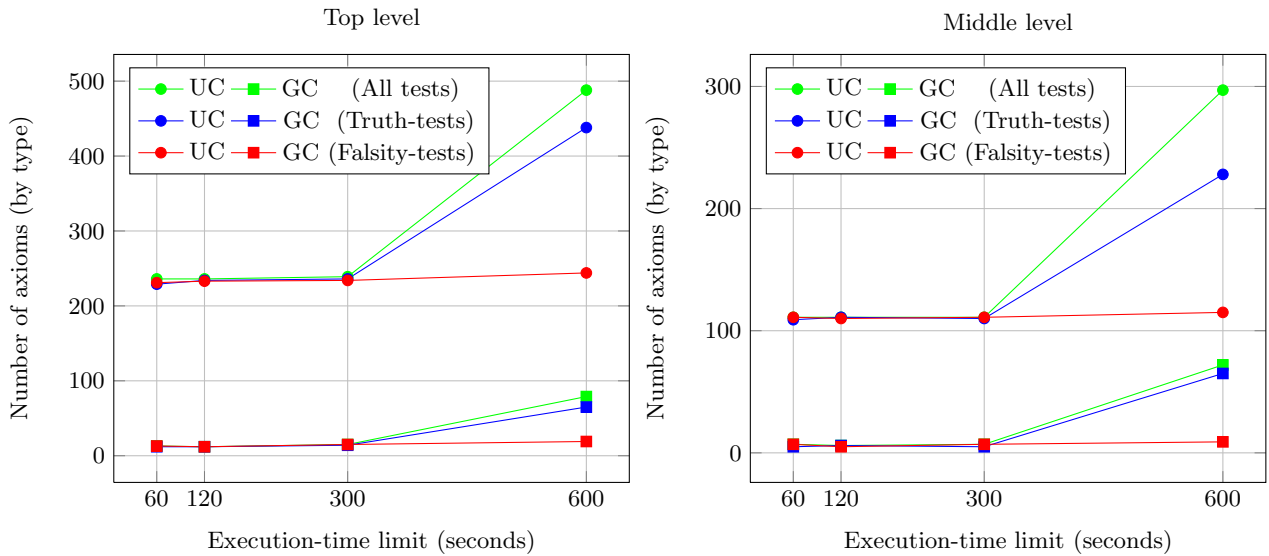
\begin{figure*}
\centering
\begin{tikzpicture}
\begin{axis}[
title=Top level,
legend columns = 2,
legend pos = north west,
xlabel = {Execution-time limit (seconds)},
ylabel = {Number of axioms (by type)},
xtick=data,
grid=major
]
\addplot[color=green,mark=*] coordinates {
	(60,236)
	(120,236)
	(300,239)
	(600,488)
};
\addplot[color=green,mark=square*] coordinates {
	(60,13)
	(120,12)
	(300,15)
	(600,79)
};
\addplot[color=blue,mark=*] coordinates {
	(60,229)
	(120,234)
	(300,236)
	(600,438)
};
\addplot[color=blue,mark=square*] coordinates {
	(60,12)
	(120,12)
	(300,14)
	(600,65)
};
\addplot[color=red,mark=*] coordinates {
	(60,231)
	(120,233)
	(300,234)
	(600,244)
};
\addplot[color=red,mark=square*] coordinates {
	(60,13)
	(120,12)
	(300,15)
	(600,19)
};
\legend{ UC, GC ~~~(All tests)~~~, UC, GC ~(Truth-tests)~, UC, GC (Falsity-tests) }
\end{axis}
\end{tikzpicture}
%
\begin{tikzpicture}
\begin{axis}[
title=Middle level,
legend columns = 2,
legend pos = north west,
xlabel = {Execution-time limit (seconds)},
ylabel = {Number of axioms (by type)},
xtick=data,
grid=major
]
\addplot[color=green,mark=*] coordinates {
	(60,111)
	(120,111)
	(300,111)
	(600,297)
};
\addplot[color=green,mark=square*] coordinates {
	(60,7)
	(120,6)
	(300,7)
	(600,72)
};
\addplot[color=blue,mark=*] coordinates {
	(60,109)
	(120,111)
	(300,110)
	(600,228)
};
\addplot[color=blue,mark=square*] coordinates {
	(60,5)
	(120,6)
	(300,5)
	(600,65)
};
\addplot[color=red,mark=*] coordinates {
	(60,111)
	(120,110)
	(300,111)
	(600,115)
};
\addplot[color=red,mark=square*] coordinates {
	(60,7)
	(120,5)
	(300,7)
	(600,9)
};
\legend{ UC, GC ~~~(All tests)~~~, UC, GC ~(Truth-tests)~, UC, GC (Falsity-tests) }
\end{axis}
\end{tikzpicture}
\caption{Number of axioms from the top and middle levels}
\label{fig:TotalAxiomsTopMiddleLevel}
\end{figure*}

In order to provide a complete evaluation framework, in \cite{ALR15} the authors also propose a set of CQs as benchmark. The proposed CQs are semi-automatically obtained from the mapping from \WORDNET to \SUMO \cite{Niles+Pease'03} and the information in the morphosemantic database,\footnote{Available at \url{http://wordnetcode.princeton.edu/standoff-files/morphosemantic-links.xls}.} which contains semantic relations between morphologically related nouns and verbs. More concretely, the authors generate 3,556 truth-tests that are obtained from 5 patterns defined on the basis on the information about antonyms and process. By negating the conjectures in the resulting truth-tests, the authors also propose 3,556 falsity tests. Consequently, the proposed benchmark consists of 7,112 CQs. This benchmark is used to evaluate \ADIMENSUMO v2.4, and the results are described in the next section.

\section{Evaluation results} \label{section:Evaluation}

In this section, we report the results of the evaluation of \ADIMENSUMO v2.4 using the framework introduced in the previous section. For this evaluation, we have used the ATP Vampire 3.0\footnote{\url{http://www.vprover.org}} \cite{RiV02} in a standard 64-bit Intel\textregistered~Core\texttrademark~i7-2600 CPU @ 3.40GHz desktop machine with 16GB of RAM. We have considered 4 different execution-time limits: 60, 120, 300 and 600 seconds. The ontology \ADIMENSUMO v2.4, the set of CQs and the execution reports are freely available.\footnote{\url{http://adimen.si.ehu.es/web/AdimenSUMO}}

In Figure \ref{fig:SolvedGoals}, we sum up the number of solved goals within different execution-time limits. It is worth to mention that these solved goals include both passing truth-tests and non-passing falsity-tests. As expected, the ATP solves more problems within an execution time-limit of 600 seconds: 1,281 proofs (\%18 of 7,112 CQs), from which 894 proofs correspond to passing truth-tests (\%25 of 3,556 truth-tests) and 487 proofs corresponds to non-passing falsity-tests (\%14 of 3,556 falsity-tests). The number of problems solved within 60 seconds, 120 seconds and 300 seconds are 960, 964 and 968 respectively. However, the most outstanding result is that the number of solved goals that correspond to falsity-tests (that is, non-passing falsity-tests) does not significantly increase with the execution-time limit: 482, 482, 484 and 487 non-passing falsity-tests within 60, 120, 300 and 600 seconds respectively. This implies that the ATP solves more and more complex truth-tests as the execution-time limit becomes longer, whereas the proofs of non-passing falsity-tests are very simple and can be detected within really short execution-time limit. This conclusion is confirmed by the results reported in Figures \ref{fig:TotalAxioms} and \ref{fig:AverageAxioms} regarding the number of used axioms: in Figure \ref{fig:TotalAxioms}, we sum up the number of different axioms used in all the proofs (excluding the conjectures) and, in Figure \ref{fig:AverageAxioms}, we sum up the average number of different axioms (excluding the conjecture) used in proofs. As before, the number of different axioms used in non-passing falsity-tests within 60, 120, 300 and 600 seconds is similar: 374, 371, 380 and 401 different axioms respectively. By contrast, 831 different axioms are used in passing truth-tests within 600 seconds, whereas 365, 374 and 377 different axioms are used within 60, 120 and 300 seconds respectively. In the same way, more than 14 different axioms (excluding the conjecture) is used in the proof of each passing truth-tests within 600 seconds, whereas around 3 different axioms (excluding the conjecture) are used in all the remaining cases. To sum up, it seems that solving the goals corresponding to non-passing falsity-tests requires less knowledge than solving the goals corresponding passing truth-tests: the number of proved truth-tests increases along the execution time-limit, whereas the number of proved falsity-tests does not. Thus, it is clear that \ADIMENSUMO v2.4 provides the knowledge for solving the goals corresponding to complex truth-tests, but state-of-the-art ATPs require longer execution time-limit. On the contrary, it may happen that either (a) the goals of the remaining falsity-tests are not provable, (b) the ontology does not contain the knowledge for solving additional falsity-test goals, or (c) state-of-the-art ATPs require much longer execution time-limit for solving falsity-test goals.

Regarding the type of axioms and the knowledge used in proofs, in Figure \ref{fig:TotalAxiomsTopMiddleLevel} we sum up the number of axioms from the top level and the middle of \SUMO used in proofs, distinguishing between atomic formulas ---{\it unit clauses} (UC)--- and non-atomic formulas ---{\it general clauses} (GC)---. Around 240 unit clauses (\%12 of 2,034 unit clauses) and 15 general clauses (\%2 of 650 general clauses) from the top level of \SUMO are used in the proofs of the non-passing falsity-tests within all the considered execution-time limits, whereas 438 unit clauses (\%21 of 2,034 unit clauses) and 65 general clauses (\%10 of 650 general clauses) are used in the 894 passing truth-tests within an execution-time limit of 600 seconds. Regarding the middle level of \SUMOWWS, around 110 unit clauses (\%4 of 2,601 unit clauses) and 8 general clauses (\%1 of 975 general clauses) are used in the proofs of non-passing falsity-tests within all the considered execution-time limits, and 228 unit clauses (\%9 of 2,601 unit clauses) and 65 general clauses (\%7 of 975 general clauses) are used in the 894 passing truth-tests within an execution-time limit of 600 seconds. To sum up, 503 top level axioms (\%19 of 2,654 axioms) and 293 middle level axioms (\%8 of 3,576 axioms) are already used in the proofs of the 894 passing truth-tests. Since the CQs have been semi-automatically constructed, we consider that the ratio between the number of used axioms and the number of proofs corresponding to passing truth-tests is really high. 

Clearly, the number of different top level axioms used in proofs is higher than the number of different middle level axioms. But the difference is even bigger when considering the average number of different axioms from the top and middle levels used in the proof of each passing truth-tests: around 8 unit clauses and 1 general clause from the top level, and 0.5 unit clauses and 0.15 general clauses from the middle level. This result is not surprising since the knowledge in the middle level of \SUMO is defined on the basis of the knowledge in the top level.

\section{Concluding remarks} \label{section:ConcludingRemarks}

Our experimental results prove that the application of the methodology proposed in \cite{GrF95} to FO ontologies can be successfully automatized by using FOL ATPs. In particular, it is encouraging that state-of-the-art FOL ATPs are able to construct complex proofs for solving CQs within relatively small execution-time limits.

Additionally, these results also enable to evaluate the usefulness of axioms by checking its use when solving CQs. Regarding the proofs obtained within an execution time-limit of 600 seconds, 971 different axioms have been used (\%13 of 7,420 axioms), from which 785 are unit clauses (\%17 of 4,635 unit clauses) and 186 are general clauses (\%7 of 2,785 general clauses). Consequently, 6,449 axioms (\%87 of the total) have not been used in any proof. Among the axioms that have been used in some proofs, 154 axioms (\%16 of 971 axioms) have been used in 10 or more proofs and 127 axioms (\%13 of 971 axioms) have been used in the proof of 10 or more goals corresponding to truth-tests. This implies that these set of 127 axioms is really useful for passing truth-tests and seems to be correct. Further, since the remaining 817 axioms (\%84 of 971 axioms) are used in less than 10 proofs, we can also conclude that the quality of proposed set of CQs in \cite{ALR15} is high because the proofs involve different sets of axioms.



In future work, we plan to improve the coverage of the set of CQs to the knowledge of the ontology, in particular to the knowledge in the middle level of \SUMOWWS, by developing new methods of obtaining CQs in a semi-automatic way.

\section{Acknowledgments}

We are grateful to the anonymous reviewers for their insightful comments. This work has been partially funded by the Spanish Projects SKaTer (TIN2012-38584-C06-02) and COMMAS\; (TIN2013-46181-C2-2-R),\; the\; Basque\; Project LoRea (GIU12/26) and grant BAILab (UFI11/45).

\bibliographystyle{abbrv}
\bibliography{alr15}

\end{document}